\pdfoutput=1
\documentclass[letterpaper, 10 pt, conference, hyphens]{ieeeconf}  

\IEEEoverridecommandlockouts                              
\usepackage[utf8]{inputenc}
\usepackage[T1]{fontenc}
\usepackage{amsmath}
\usepackage{amssymb}
\usepackage{bm}
\usepackage{epsfig}
\usepackage{graphics}
\usepackage{mathptmx}
\usepackage{hyperref}
\usepackage{mathtools}
\usepackage{xcolor}

\title{\LARGE \bf Quantum Optical Convolutional Neural Network:
A Novel Image Recognition Framework for Quantum Computing*
}

\author{Rishab Parthasarathy$^{1}$, Rohan Bhowmik$^{1}$
\thanks{*This work was not supported by any organization}
\thanks{$^{1}$Both R. Parthasarathy and R. Bhowmik are students
        at the Harker School, 500 Saratoga Ave, San Jose CA, USA 95138.
        {\tt\small (22rishabp/23rohanb)@students.harker.org}}
}

\begin{document}

\maketitle
\thispagestyle{empty}
\pagestyle{empty}

\begin{abstract}
  
Large machine learning models based on Convolutional Neural Networks (CNNs) with rapidly increasing number of parameters, trained with massive amounts of data, are being deployed in a wide array of computer vision tasks from self-driving cars to medical imaging. The insatiable demand for computing resources required to train these models is fast outpacing the advancement of classical computing hardware, and new frameworks including Optical Neural Networks (ONNs) and quantum computing are being explored as future alternatives.

In this work, we report a novel quantum computing based deep learning model, the Quantum Optical Convolutional Neural Network (QOCNN), to alleviate the computational bottleneck in future computer vision applications. Using the popular MNIST dataset, we have benchmarked this new architecture against a traditional CNN based on the seminal LeNet model. We have also compared the performance with previously reported ONNs, namely the GridNet and ComplexNet, as well as a Quantum Optical Neural Network (QONN) that we built by combining the ComplexNet with quantum based sinusoidal nonlinearities. In essence, our work extends the prior research on QONN by adding quantum convolution and pooling layers preceding it.

We have evaluated all the models by determining their accuracies, confusion matrices, Receiver Operating Characteristic (ROC) curves, and Matthews Correlation Coefficients. The performance of the models were similar overall, and the ROC curves indicated that the new QOCNN model is robust. Finally, we estimated the gains in computational efficiencies from executing this novel framework on a quantum computer. We conclude that switching to a quantum computing based approach to deep learning may result in comparable accuracies to classical models, while achieving unprecedented boosts in computational performances and drastic reduction in power consumption.

\end{abstract}

\section{\textbf{INTRODUCTION}}
\subsection{Background}
Artificial Intelligence (AI) is having an unprecedented surge in applications in the recent years based on breakthrough advances in machine learning techniques. The field of computer vision has emerged to be at the forefront of many such developments. Computer vision has been deployed in myriad applications ranging from self-driving cars to the medical fields of radiology in diagnosis. These applications often leverage machine learning frameworks based on Deep Neural Networks (DNN) trained with enormous volumes of image data, often referred to as "deep learning", to produce previously unattainable results \hyperref[c1]{\cite{c1}}.

However, AI applications, especially those built with deep learning, are consuming immense amounts of computing resources. In fact, the computational cost for AI is on a rapid rise, as OpenAI reports that the computational power required to train modern AI models is doubling every 3.5 months \hyperref[c2]{\cite{c2}}. For example, the recently introduced GPT-3 model, which is at the cutting edge of Natural Language Processing (NLP), includes a whopping 175 billion parameters, far more than had ever been reported before \hyperref[c3]{\cite{c3}}. In contrast, the number of transistors in a microprocessor chip, a traditional indicator for the historical advances in the processing capability of general-purpose computing hardware, has been doubling about every two years, per Moore’s Law. Recently, even this two-year doubling capability of classical computing hardware has been called into doubt, with the common idea of shrinking transistors no longer being as effective or possible \hyperref[c4]{\cite{c4}}. In order to alleviate this bottleneck, AI engineers have been increasingly switching to massively parallel computing architectures such as the Graphical Processing Units (GPU) or custom-built AI accelerator chips \hyperref[c5]{\cite{c5}}\hyperref[c6]{\cite{c6}}. Despite these trends, the insatiable demand for computing resources continue to rapidly outpace the progress in hardware capabilities. The fast-increasing computational needs of AI will inevitably reach a breaking point that will risk limiting progress. Therefore, we must address the pressing need for AI computing resources with new architectures and algorithms. 

Quantum computing is one promising field that can potentially help address this problem with radically different architectures. Therefore, research into new deep learning techniques based on quantum computers is imperative to attempt to keep up with AI breakthroughs in the future. Quantum computing is a completely new paradigm for computers that involves quantum physics instead of classical physics. Instead of the classical bits, quantum computing involves quantum bits, or qubits for short, that operate with superposition and the uncertainty characteristic of quantum physics \hyperref[c7]{\cite{c7}}. Quantum algorithms have been shown to be capable of achieving exponential gains in comparison to classical algorithms in many fields, including nondeterministic polynomial time (NP) problems such as integer factorization and discrete logarithms by means of Shor's Algorithm and many more \hyperref[c8]{\cite{c8}}\hyperref[c9]{\cite{c9}}. In these cases, problems that cannot be feasibly solved classically can be solved by a quantum computer in a reasonable time scale. In addition, many AI operations utilize linear algebra, especially processes like matrix multiplication and eigendecomposition, and quantum computers have been proven to theoretically have a significant advantage over classical computing in these operations \hyperref[c10]{\cite{c10}}\hyperref[c11]{\cite{c11}}. Also, quantum volume, a measure of the amount of data that current quantum computers can store, is rapidly increasing \hyperref[c12]{\cite{c12}}. For these reasons, research into quantum AI is picking up pace in the recent years.

\subsection{Prior Research}

Recently, much research has been conducted in order to decrease the computational costs of training large AI models through the transfer learning techniques, in which models are created with pretrained parameters on large datasets and only the final layers of the model are retrained with new data for the target applications. Essentially, transfer learning looks for a way to effectively transfer data between models to speed up training the AI models for new applications \hyperref[c13]{\cite{c13}}.

In order to achieve significantly higher performances, researchers have explored and reported the feasibility of extending machine learning frameworks into the quantum space \hyperref[c14]{\cite{c14}}. Prior research has also verified that quantum machine learning algorithms have significant potential in performing the classical tasks \hyperref[c15]{\cite{c15}}.

In addition, several researchers have developed different formats of Optical Neural Networks (ONN), including the diffraction based and interferometer based networks \hyperref[c16]{\cite{c16}}\hyperref[c17]{\cite{c17}}. In particular, the paper by Fang has reported successful implementation of ONNs utilizing Mach-Zehnder Interferometers (MZI). These networks encoded the linear transformations typical of perceptrons through differently structured arrays of MZIs. We have used the models and codes used in this paper for benchmarking purposes, and built further on the structure reported \hyperref[c17]{\cite{c17}}\hyperref[c18]{\cite{c18}}.

There has also been research into Quantum Optical Neural Networks (QONN). These networks build on the ONN structure by introducing multi-particle unitaries and quantum states in order to facilitate the usage of quantum frameworks \hyperref[c10]{\cite{c10}}. Researchers have also developed Quantum Convolutional Neural Networks (QCNN), which involve taking the convolutions and pooling typical of classical Convolutional Neural Networks (CNN) and translating them into quantum form \hyperref[c19]{\cite{c19}}.

\subsection{Overview}

In this work, we have architected and built a novel Quantum Optical Convolutional Neural Network (QOCNN) that operates as a fusion of both the quantum and optical paradigms. Specifically, we combine the preexisting capabilities of quantum computers to encode a convolution like operation with the capabilities of quantum photonics that yield linear transformations and nonlinearities. Using this methodology, We demonstrate that the new QOCNN architecture, simulated on a classical computer, can achieve accuracies that are comparable to CNN, ONN, and QONN models, while offering substantial gains in computational efficiencies for execution on the future quantum computing hardware.

\section{\textbf{METHODS}}

\subsection{Dataset}

In order to train and evaluate our proposed QOCNN framework, as well as various other AI models for benchmarking purposes, we used the MNIST dataset, which consists of 28X28 grayscale images of handwritten digits \hyperref[c20]{\cite{c20}}\hyperref[c21]{\cite{c21}}\hyperref[c22]{\cite{c22}}. We split the 70,000 images available in the dataset, allocating 60,000 images for training and 10,000 images for testing the AI models. The 10,000 testing images were sequestered in order to avoid overfitting the models.

We pre-processed the MNIST image data for use with the quantum models by superimposing the top half of the image over the bottom half through a translation step in order to generate 392 complex numbers, which are also single photon Fock states, as will be explained later \hyperref[c17]{\cite{c17}}. Specifically, if $a$ is the matrix of pixel values, then $a_{x, y}$ is the real part and $a_{x + 14, y}$ is the complex part of each complex number for $x \leq 14$ and $y \leq 28$. To process this dataset classically, we flattened the array into one dimension. In essence, each input matrix $I$ is represented as $[I_{\text{re}}, I_{\text{im}}]$, where $I_{\text{re}}$ represents the real part of the input complex numbers and $I_{\text{im}}$ represents the imaginary part.

This means that 392 qubits in a quantum computer hardware can run this network, which should be feasible in the foreseeable future with the continued pace of progress in quantum computing hardware.

\subsection{Convolutional Neural Network (CNN)}

\begin{figure}[thpb]
      \centering
      \label{cnnstructure}
      \includegraphics[width = 3.5in]{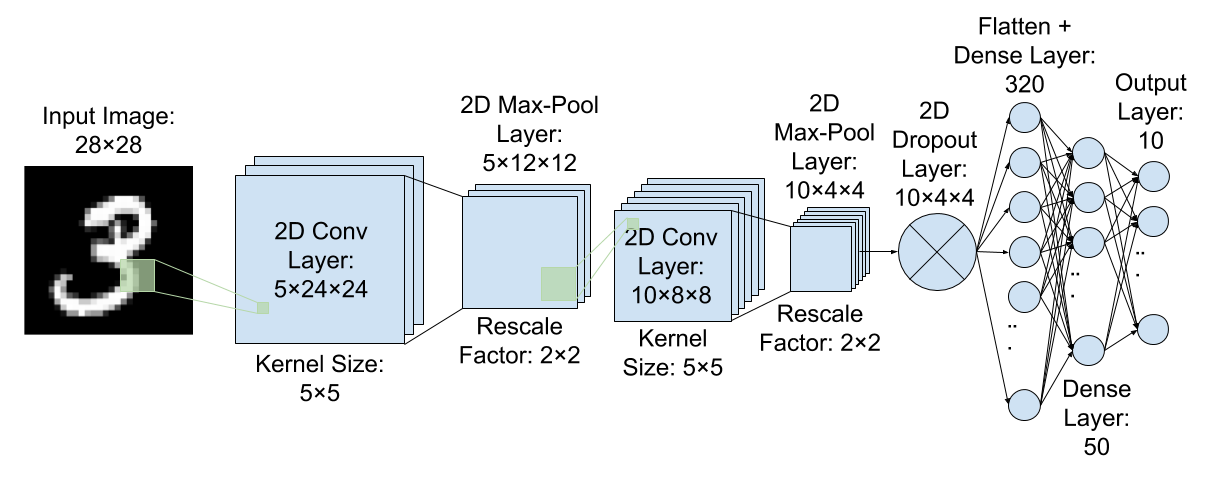}
      \caption{This figure depicts the structure of our benchmark convolutional neural network, which is a variation of LeNet: two 2D Convolutional and 2D Max-Pooling Layers, followed by a 2D Dropout, a Flatten, and two Dense Layers to produce an Output Layer of ten neurons. As a traditional neural network, this architecture takes in real numbers as inputs, representing image pixel values and uses Rectified Linear Unit (ReLU) as its activation function. $P_\alpha$, the value of output neuron $\alpha$, represents the probability for digit $\alpha$.}
\end{figure}

In order to benchmark our model against the current standard, we developed a classical convolutional neural network in PyTorch that is a variation of the famous LeNet model used on MNIST \hyperref[c23]{\cite{c23}}, as shown in \hyperref[cnnstructure]{Fig. 1}. 

As shown in \hyperref[cnnstructure]{Fig. 1}, our CNN architecture is a variation of the LeNet structure, consisting of groupings of Convolutional Layers of 5x5 kernels and 2x2 Maximum Pooling Layers followed by Rectified Linear Unit (ReLU) nonlinearity. Then, we incorporated a Dropout Layer and flattened the resulting data into one dimension before feeding them into two Fully Connected (Dense) Layers to reduce the number of neurons to 10. Finally, we applied a Log Softmax transformation, which normalized the values of the output neurons so that the value $P_\alpha$ represented the probability of the image representing the digit $\alpha$. This essentially means that the the values of the output neurons $P_\alpha$ satisfy $\sum_{\alpha = 0}^9 P_\alpha = 1$.

This network was trained with the 60,000 training images from the MNIST dataset for 10 epochs or until the test curve began to flatten in order to prevent overfitting.

\subsection{Optical Neural Networks (ONN)}

As another standard for benchmarking comparisons, we used the optical neural networks presented by Fang et al. \hyperref[c17]{\cite{c17}}. We used the libraries published in this work as the basis for our model and implemented them in the PyTorch framework \hyperref[c18]{\cite{c18}}.

Specifically, the optical neural networks modified the multilayer perceptron by keeping the similar structure of matrix multiplication, but using Singular Value Decomposition (SVD), which is described in \hyperref[svd]{Equation 1}.

\begin{equation}
    M = V\Sigma U.
    \label{svd}
\end{equation}

\begin{figure}[thpb]
      \centering
      \includegraphics[width = 3.3in]{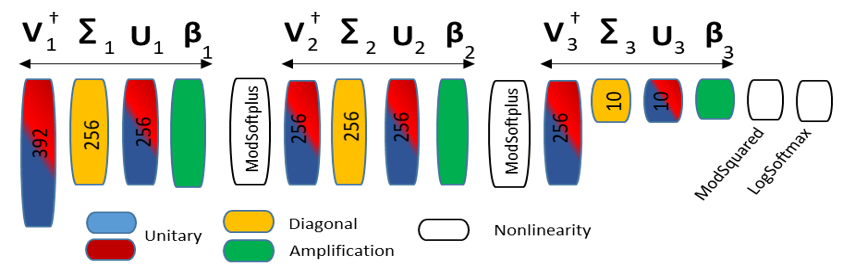}
      \label{onnstructure}
      \caption{We illustrate the typical structure of a simple three-layer ONN. Each layer contains the three parts of a SVD decomposed linear matrix, which are two unitary matrices separated by a diagonal matrix, and an amplification to enable precise MZI behavior for the unitaries. In the figure, the first unitary, diagonal, second unitary, and amplification in each layer are denoted by $V$, $\Sigma$, $U$, and $\beta$, respectively. Finally, the layers before the end are capped of with a ModSoftplus layer, which is a discrete sigmoidal nonlinearity, and the last layer is ended with a ModSquared discrete nonlinearity and a LogSoftmax, which is a smooth sigmoid that produces the final predictions \hyperref[c17]{\cite{c17}}.}
\end{figure}

As depicted in \hyperref[onnstructure]{Fig. 2} along with \hyperref[svd]{Eq. 1}, each linear matrix transformation can be decomposed into three parts: two unitaries $V$ and $U$ and a diagonal matrix $\Sigma$ \hyperref[c24]{\cite{c24}}. This operation can be visualized in a standard formalism as a sequence consisting of a rotation, an independent scaling on each basis vector, and then a final rotation.

Then, in order to ensure precision by the Mach-Zehnder Interferometers (MZIs) in the diagonal because each value can be at most 1, an amplification layer is included at the end that scales the values. Thus, \hyperref[svd]{Equation 1} becomes 
\begin{equation}
    M = V \Sigma U \beta.
    \label{onnsvd}
\end{equation}

In this form, $\beta$ serves to uniformly scale the output to make up for the small values in $\Sigma$. Then, the hidden layers end with a discrete sigmoid nonlinearity, while the final layer ends with a discrete nonlinearity followed by the typical sigmoidal prediction phase of a CNN, the Softmax.

For benchmarking the ONN, we used a pretrained GridNet (from the original paper by Fang et al.) along with a ComplexNet (optimal behavior of the SVD also from the original paper) that we trained from scratch \hyperref[c17]{\cite{c17}}. 

Along with the preexisting networks, we added new code functionalities for quick extraction of data from the networks, which includes network structure, confusion matrices, and more.

\subsection{Quantum Optical Neural Networks (QONNs)} 

Building on the ONNs, we proceeded to quantum frameworks and implemented our own QONNs based on the formalism presented by Steinbrecher et al. \hyperref[c10]{\cite{c10}}. 

\begin{figure}[thpb]
      \centering
      \label{qonnstructure}
      \includegraphics[width = 3in]{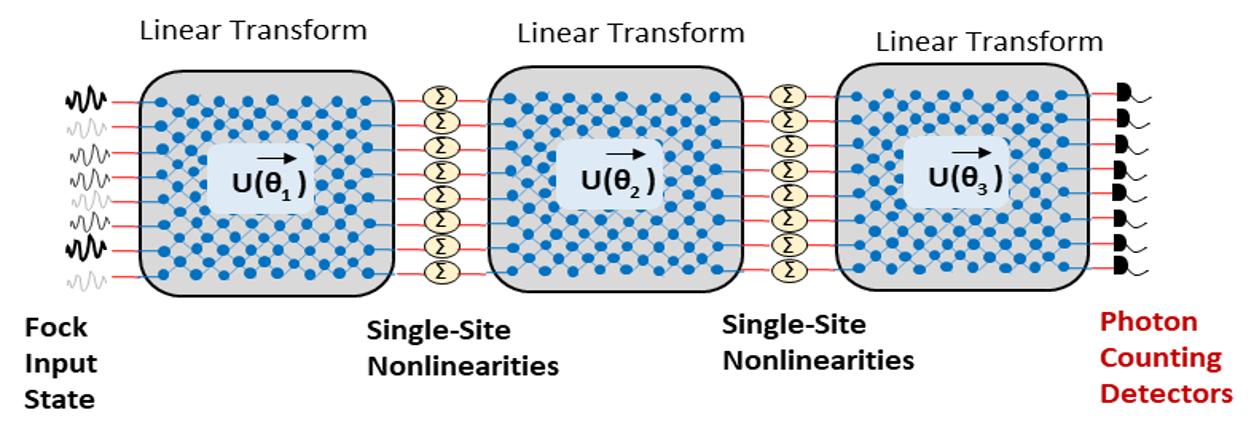}
      \caption{The QONN structure starts out by accepting Fock input states, which essentially denote groups of particles. In our case, we opted for simple single-photon Fock states, which are essentially complex numbers. Then, these Fock states are passed through linear transformations denoted by $U(\vec{\theta})$ and then individually passed through nonlinearities denoted by $\Sigma$. Specifically, in the QONN, these nonlinearities are sinusoidal \hyperref[c10]{\cite{c10}}.}
\end{figure}

As depicted in \hyperref[qonnstructure]{Fig. 3}, the main differences between the ONN and the QONN were that the QONN had quantum Fock states instead of photons as the inputs and also used multiparticle unitary matrices along with quantum-based single-site nonlinearities \hyperref[c10]{\cite{c10}}.

In addition, the QONN framework was suitable for computer vision tasks because the built-in nonlinearities turned out to support sinusoidal modes, which we then implemented on the ONN \hyperref[c25]{\cite{c25}}. 

Specifically, each nonlinearity $\Sigma$ follows the equation

\begin{equation}
    \Sigma (x) = x\sin{(\lambda x).}
    \label{nonlin}
\end{equation}

As described in \hyperref[nonlin]{Equation 3}, the nonlinearity allowed us to scale each term independently of the other terms by using a sinusoidal function. Also, to adjust the flatness and distribution of the sinusoid, a scaling hyperparameter $\lambda$ allows for modification and optimization. In this case, we found that $\lambda = 0.2$ was optimal.

Then, in order to make sure that our modification of the ONN into a QONN would be able to function with relatively large amounts of data without requiring enormous classical computing power or access to sufficiently large or error-tolerant quantum computers, we made a few key modifications to the QONN design.

First, we found that just like the ONN, the QONN could perform computer vision tasks with single-photon Fock states, or complex numbers, as inputs which significantly simplified the intermediate computations. Instead of a vector $\vec{\varphi_\alpha}$ of different particle states for each input $\alpha$, we solely have a complex number $\theta_\alpha$ for each input $\alpha$. Thus, when we create our matrix, instead of creating a matrix transformation for a vector of vectors $\vec{\Phi}$, we just create a traditional matrix transformation $U(\vec{\Theta})$ for a vector $\vec{\Theta}$.

Secondly, we realized that simulating pure unitaries would be unfeasible with the time and access to computing power that we had because all unitary optimization algorithms that we had access to utilized permanent calculation, which is a \#P-hard problem. Instead, we used linear matrices, which are not purely unitaries, but can also be executed on quantum computers in the future with more fault-tolerant NISQ qubits that allow the probabilistic execution of non-unitary operations \hyperref[c26]{\cite{c26}}\hyperref[c27]{\cite{c27}}. 

Each of these complex matrix transformations $M$ was of size $N_1 \times N_2$, where $N_1$ was the input size and $N_2$ was the output size. However, to facilitate coding, we separated $M$ into $M_R$ and $M_C$, the real and complex parts of the matrix, respectively. Then, we performed a linear transformation as defined in the following equation:
\begin{equation}
    M_F = \begin{bmatrix} M_R & M_C \\ -M_C & M_R \end{bmatrix},
    \label{complexdecomp}
\end{equation}
where $M_F$ is the final linear transformation. Because the input is stored in the one dimensional matrix $I_F = \begin{bmatrix} I_R & I_C \end{bmatrix}$, the transformation $I_FM_F$ produces $I_{1, F} = I_FM_F = \begin{bmatrix} (IM)_R & (IM)_C\end{bmatrix}$, as expected.

With these modifications, we were able to add on to the ComplexNet described by Steinbrecher et al. to build our own QONN.

\subsection{Quantum Optical Convolutional Neural Network (QOCNN)}

\begin{figure}[thpb]
      \centering
      \includegraphics[width = 3.3in]{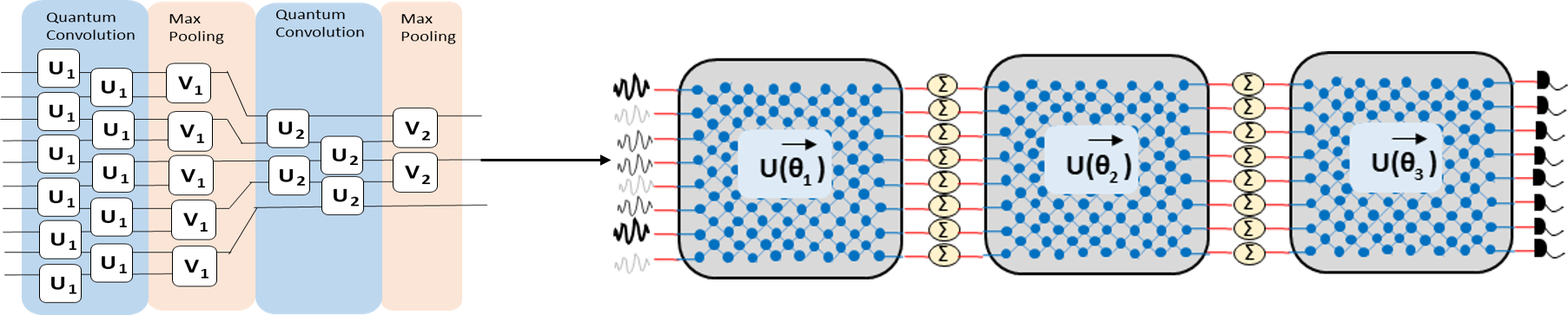}
      \label{qocnnstructure}
      \caption{This figure presents the structure of our proposed Quantum Optical Convolutional Neural Network (QOCNN). The left half of the figure depicts the quantum convolutional part of the network. Specifically, the blue layers describe the translationally invariant matrix-multiplication based quantum convolution, with $U$ representing the convolution applied. Then, the red layers represent the Max Pooling that was applied, where each $V$ represents one region that is pooled into one value. At the end of the left half, the bold arrow represents the output quantum data from the quantum convolution part of the network being passed onto the QONN part of the network, similar to that depicted in \hyperref[qonnstructure]{Fig. 3}, with linear transformations $U(\vec{\theta})$ and nonlinearities $\Sigma$.}
   \end{figure}

Finally, we created a novel Quantum Optical Convolutional Neural Network (QOCNN) that combines the Quantum Convolutional Neural Network and the Quantum Optical Neural Network frameworks. In essence, the QOCNN takes the structure of a QONN and adds a quantum convolution and pooling operations, as depicted in \hyperref[qocnnstructure]{Fig 4.}

Technically, the precise functionality of the convolution cannot be performed on a quantum computer, but there is an alternative approach \hyperref[c28]{\cite{c28}}. As proposed by Cong et al., translationally invariant matrix multiplications possess the same end-properties as the traditional convolutions \hyperref[c19]{\cite{c19}}. As depicted in \hyperref[qocnnstructure]{Fig. 4}, the same matrix is used to transform various sections of the input data, modeling the behavior of a typical classical convolution.
  
Similarly, we also deploy a translationally invariant pooling operation. Finally, at the end of the network, we use our QONN framework just like a fully-connected layer is used in a traditional CNN to extract information from the features resulting from the convolutions, as shown in \hyperref[qocnnstructure]{Fig. 4.}

Thus, the QOCNN we propose is a completely new architecture that integrates two state of the art techniques, the quantum paradigm and the optical paradigm, for a novel approach to executing computer vision tasks on the future quantum computing hardware. However, in order to make this new framework suitable for execution on classical computers for simulation and benchmarking purposes, we made certain modifications.

For example, as we explained before, we modified the matrix transformations in the quantum convolutions to linear matrices from unitary matrices in order to circumvent \#P-hard problems in backpropagation and optimization \hyperref[c26]{\cite{c26}}\hyperref[c27]{\cite{c27}}.

For the convolution operations, we first defined a convolutional kernel $K$ of size $k\times k$ and a stepsize of $s$ for an input of size $D$. Then, we defined the total number of matrix multiplications needed as $n = \lceil \frac{k}{s} \rceil$. This is because we want to superimpose the different kernels through matrix multiplication, not addition, in order to simulate the behavior of multiple quantum gates. We could superimpose nonoverlapping kernels in the same matrix without compromising matrix multiplication, but could not superimpose overlapping ones. Thus, we would superimpose the closest convolutional kernels possible in each of the $n$ matrices created for matrix multiplication.

In order to create the matrices for matrix multiplication, we defined the effective stepsize $s_0$ in each matrix used for matrix multiplication as the distance between the two closest kernels that could be superimposed without overlapping, which satifies $s_0 = n*s - k$. The maximal number of convolutional kernels that could be superimposed in the first matrix for the multiplication was also defined as 
\begin{equation}
    k_{\text{tot}} = \left\lceil \frac{D}{k + s_0} \right\rceil = \left\lceil \frac{D}{n\cdot s} \right\rceil.
    \label{kernels}
\end{equation}

As depicted in \hyperref[kernels]{Eq. 5}, each kernel and the subsequent gap takes up $k + s_0 = n\cdot s$ size, and the ceiling was taken because there could be a partial kernel in the superimposed matrix. 

We defined the $n$ matrices for multiplication as $M_1 \dots M_n$ and the final convolution as $M_f = M_1 M_2 \dots M_n$. Once again, the matrices were established in the format 
\begin{equation}
    M = \begin{bmatrix} M_R & M_C \\ -M_C & M_R \end{bmatrix},
    \label{matrixdecomp}
\end{equation}
where $M_R$ and $M_C$ are the real and imaginary parts, respectively, of the complex numbers in $M$. However, in this section, we will simply refer to the complex numbers contained within $M$ as $M_{i, j}$ with the expectation that both the real and imaginary parts are properly communicated.

We defined $M_1$ as a matrix of zeros apart from
\begin{equation}
    \begin{aligned}
    \forall 0 \leq x < k_{\text{tot}}, 1 \leq y \leq k, 1 \leq z \leq k, \\ x \cdot (k + s_0) + y \leq D, x \cdot (k + s_0) + z \leq D, \\ (M_1)_{x \cdot (k + s_0) + y, x \cdot (k + s_0) + z} = K_{y, z}.
    \end{aligned}
    \label{matrixdefinition}
\end{equation}

In essence, as depicted in \hyperref[matrixdefinition]{Eq. 7}, $M_1$ consisted of all zeros apart from the convolutional kernels superimposed on the top-left to bottom-right diagonal, with the first convolutional kernel beginning in the top left corner of the matrix.

Building on the definition of $M_1$, $M_i$ was defined recursively as a matrix of zeros apart from
\begin{equation}
    \forall s < x \leq D, s < y \leq D, (M_i)_{x, y} = (M_{i - 1})_{x - s, y - s},
    \label{recursivematrixdefinition}
\end{equation}
which means that all the kernels were just shifted by one stepsize. In this manner, all $M_i$ were defined and the final convolution $M_f$ was calculated as $M_1 M_2 \dots M_n$ before being applied to the input data via matrix multiplication.

Second, we applied classical max pooling operation in a translationally invariant fashion as depicted in \hyperref[qocnnstructure]{Fig. 4}, instead of a quantum pooling algorithm which would be infeasible to simulate on a classical device. In order to do so, we first separated the input data $I = [I_{\text{re}} I_{\text{im}}]$ into its real and complex parts, $I_{\text{re}}$ and $I_{\text{im}}$, respectively, because the input data for the real and complex parts represents two different parts of the image, the top and the bottom, respectively, as we described earlier. Then, we applied separate max pooling to $I_{\text{re}}$ and $I_{\text{im}}$ before once again concatenating the output into $I_F = [I_{F, \text{re}} I_{F, \text{im}}]$ in order to facilitate further processing of data.

Specifically, we utilized one convolutional layer and one max pooling layer to extract key features from the input data. Then, we applied the QONN framework as defined in the previous section and in \hyperref[qocnnstructure]{Fig. 4} to produce a prediction for the image.

\section{\textbf{RESULTS}}

\subsection{Accuracies and Advanced Statistics}

\begin{figure*}[thpb]
    \centering
    \label{accuracyanalysis}
    \includegraphics[width = 6.5in]{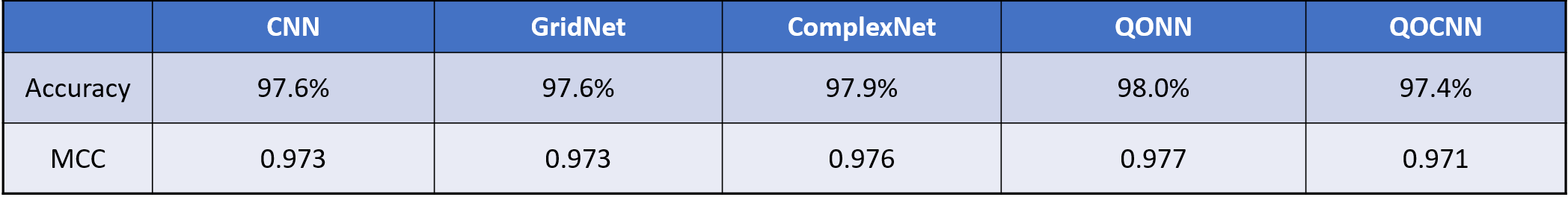}
    \caption{This table lists the accuracies and Matthews Correlation Coefficients (MCC) for all 5 Networks we evaluated: a) the classical CNN model, a variation of the LeNet \hyperref[c23]{\cite{c23}}; b) the GridNet, one of the ONN models \hyperref[c17]{\cite{c17}}; c) the ComplexNet, which is the ONN with ideal training of the Singular Value Decomposition (SVD) \hyperref[c17]{\cite{c17}}; d) the QONN that we developed with a $\sin{x}$ sinusoidal nonlinearity \hyperref[c10]{\cite{c10}}; and e) our proposed QOCNN model.}
\end{figure*}
\begin{figure*}[thpb]
      \centering
      \includegraphics[width = 6.1in]{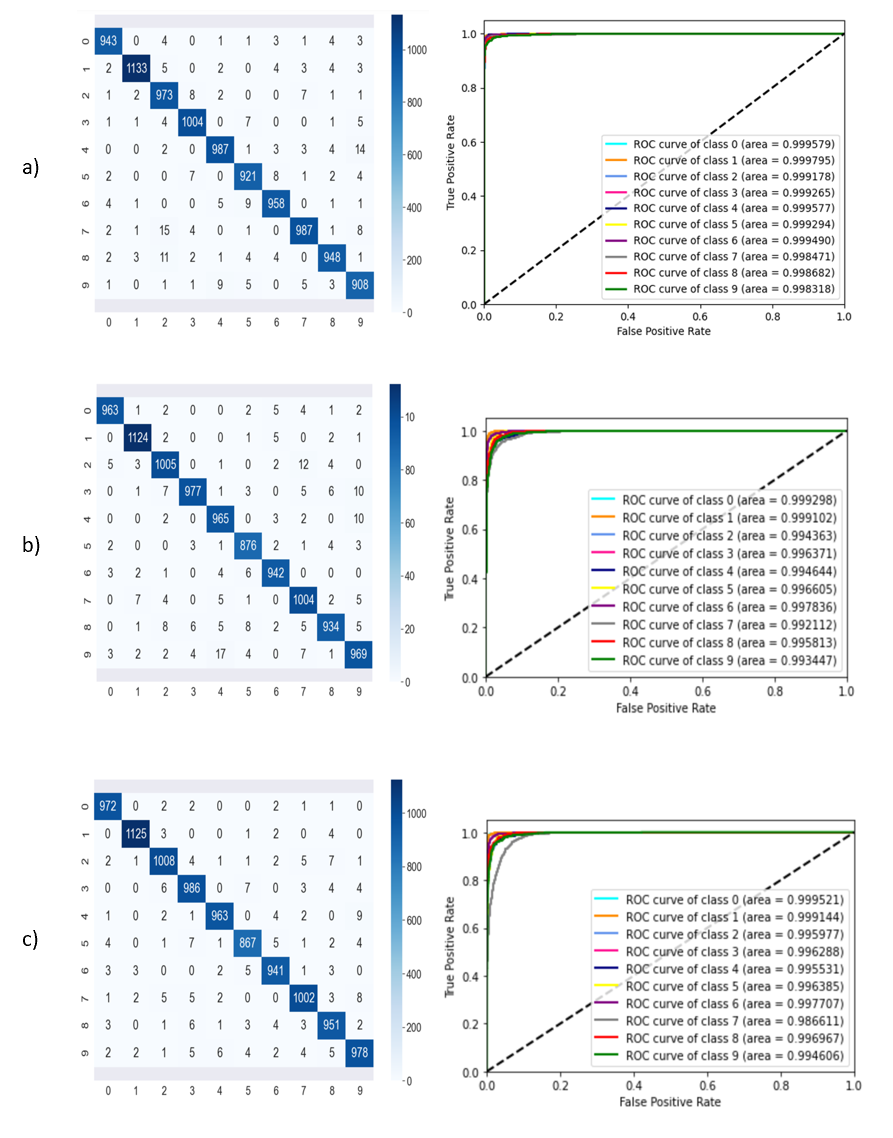}
      \label{cnnanalysis}
      \label{gridnetanalysis}
      \label{complexnetanalysis}
\end{figure*}
\begin{figure*}[thpb]
    \includegraphics[width = 6.1in]{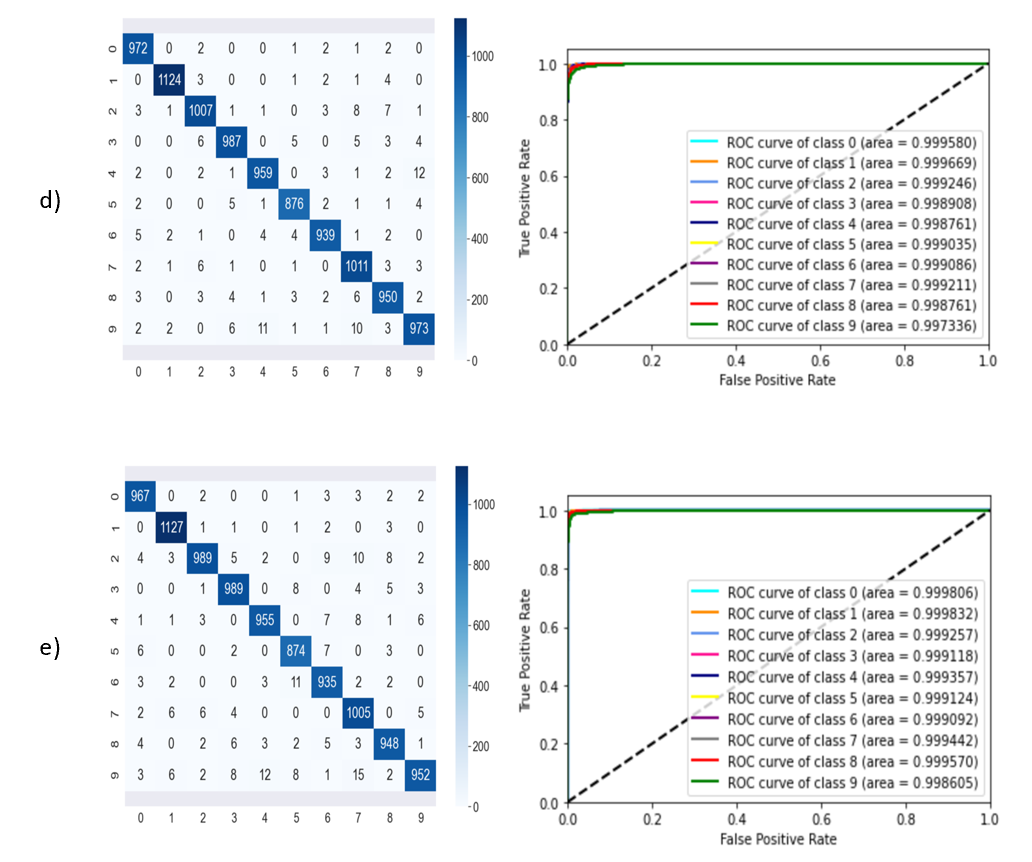}
    \label{qonnanalysis}
    \label{qocnnanalysis}
    \caption{This figure presents the confusion matrices on the left and Receiver Operating Characteristic (ROC) curves on the right in each row: (CNN, GridNet, ComplexNet, QONN with $\sin{x}$ nonlinearity, and finally the proposed QOCNN model). Confusion matrices graph the expected responses against the received responses, with the top left to bottom right diagonal being the intended location. These confusion matrices are heatmapped with white being low frequency and dark blue representing high frequency. ROC curves depict the performance of a model as the confidence level required to make a prediction is varied. The classes 0 through 9 represent the images of the numbers 0-9, respectively. Each ROC curve is for a different number, and each curve has the area under it displayed, with 1 being ideal.}
\end{figure*}

In order to evaluate the comparative performance of all five networks that we built (CNN, GridNet, ComplexNet, QONN with a $\sin{x}$ nonlinearity, and our proposed QOCNN model), we used four quantitative metrics. 

The first metric we evaluated was the accuracy of prediction, and the second was Matthews Correlation Coefficient (MCC), which is given by 
\begin{equation}
MCC = \frac{TP * TN - FP*FN}{\sqrt{(TP + FP)(TP + FN)(TN + FP)(TN + FN)}},
\end{equation}
where TP, TN, FP, and FN are true positive, true negative, false positive, and false negative, respectively. The accuracies and MCC values that we determined for all 5 networks are shown in \hyperref[accuracyanalysis]{Fig. 5}. In addition, we also utilized the confusion matrices and the Receiver Over Characteristic (ROC) curves to compare the performance of the various networks. Confusion matrices graph the expected responses against the received responses, with the top left to bottom right diagonal being the intended location. As shown in \hyperref[qocnnanalysis]{Fig. 6}, these confusion matrices are heatmapped with white being low frequency and darker blues representing higher frequencies. ROC curves depict the performance of a model as the confidence level required to make a prediction is varied. The classes 0 through 9 represent the images of the numbers 0-9, respectively. Each ROC curve shown is for a different number, and each curve has the area under it displayed, with 1 being ideal. Confusion matrices and ROC curves for all 5 networks are presented in \hyperref[cnnanalysis]{Fig. 6(a)}, \hyperref[gridnetanalysis]{Fig. 6(b)}, \hyperref[complexnetanalysis]{Fig. 6(c)}, \hyperref[qonnanalysis]{Fig. 6(d)}, and \hyperref[qocnnanalysis]{Fig. 6(e)}.
As we can see from these figures, all of the models performed quite well on the MNIST dataset. While all the models achieved comparable results, it is worth noting that unlike the confusion matrices of the other models that have a few focused hotspots, the confusion matrix of the QOCNN has a more distributed error profile, which suggests that it has not overfit and that additional training could enhance the accuracy even further.

In addition, the MCC, which quantizes the false positive and true negative rates into one statistic, also finds that the performances of all the models are relatively similar, with MCCs around 0.97. 

However, our proposed QOCNN model stands out when we compare the ROC Curves. As we can see, the ROC curves for GridNet, ComplexNet, and the QONN show some deviations from the rectangular perfect curve, indicating lack of robustness from those models in certain cases. This could mean that while these networks were learning solidly for a simple dataset like MNIST, when a more complex dataset that required significantly more detail recognition was presented, the networks may not perform as well. 

The ROC curves produced by our QOCNN model are spectacular, even surpassing the ones of the CNN in Area under Curve metric. This means that the QOCNN is learning complex information, and its ability to compete with the CNN is telling, because the primary reason for the usefulness of CNN architectures in modern computer vision research is its ability to extract significant information from images.

This robustness bodes well for the applicability of our proposed QOCNN framework to more advanced and complicated datasets, especially with the fact that its accuracy and confusion matrices are also competitive.

In addition, quantum and optical frameworks are theorized to take around $10^{7}$ times less power and achieve higher operations per second than a classical computer, which would help lessen the computational burden rather substantially. This is because the optical regime can communicate a value with a short pulse of light with energy in the attojoules, whereas the classical regime requires a longer and more extended electrical pulse with energy in the picojoules for each operation \hyperref[c29]{\cite{c29}}.

At the current moment, there is no experimental proof of the efficiency boosts that future quantum computing systems can provide over the modern classical neural networks. Especially with lossy qubits, there is no true evidence yet that quantum computing can truly boost neural network efficiencies \hyperref[c30]{\cite{c30}}. However, because our proposed QOCNN framework relies almost completely on matrix multiplications, which theoretically has a speed-up over the classical regime \hyperref[c31]{\cite{c31}}, we believe that with advances in the quantum technology that is fast emerging with the advent of NISQ qubits, these computations will become possible.

Theoretically, matrix multiplication takes less time in quantum computing than classically. This can be achieved through a procedure known as the swap test, which performs a quantum state inner product in the same speed that it takes to make the quantum states themselves, which shaves off one order of matrix size from efficiency \hyperref[c31]{\cite{c31}}. In a classical setting, a naive matrix multiplication of matrices of size $b$x$n$ and $n$x$n$, where $b$ is batch size and $n$ is input size, would have an efficiency of $\mathcal{O}\left(bn^2\right)$. However, using the inner product, this same matrix multiplication would have an efficiency of $\mathcal{O}\left(n(n + b)\right)$. This means that given some number of layers $L$, quantum hardware theoretically increases forward propagation efficiency from $\mathcal{O}\left(bn^2L\right)$ to $\mathcal{O}\left((bn + n^2)L\right)$. Similarly, quantum backpropagation would see the same increase of efficiency as forward propagation.

Similarly, the memory usage by classical frameworks is $\mathcal{O}\left(n^2L\right)$ to account for each of the approximately $n^2$ parameters in $L$ layers, but in a quantum system, there would theoretically only be $\mathcal{O}\left(L(\log{n} + \log{b})\right) = \mathcal{O}\left(n\log{\left(nb\right)}\right)$ qubits used. This is because each qubit can double the amount of data each layer can store, resulting in $\mathcal{O}\left(\log{n}\right)$ parameters per layer \hyperref[c19]{\cite{c19}}. In addition, the input data takes up $\mathcal{O}\left(\log{nb}\right)$ data for each layer, which sums to $\mathcal{O}\left(L(\log{n} + \log{b})\right)$.

For example, for a network with 10 layers, a batch size of 200, and an input size of 10,000, the efficiency would increase approximately 200 fold. In addition, the model would have $10^9$ parameters in the classical framework, but would only require around $200$ qubit parameters when executed using a quantum framework. 

Thus, large neural networks implemented in quantum frameworks would achieve an efficiency boost of multiple orders of magnitude over classical frameworks, thereby offering a significant step forward for machine learning.

Specifically, the QOCNN framework that we have developed has a tremendous potential to be a significant computationally-efficient alternative executed on future quantum computing hardware to the traditional CNN models implemented on classical computers, while achieving similar accuracies and robustness.

\section{\textbf{CONCLUSION}}

We developed and evaluated a novel machine learning algorithm for computer vision and image recognition tasks, with a goal to achieve substantial computational efficiencies on future quantum computing hardware platforms. Specifically, we proposed a quantum optical convolutional neural network, or QOCNN for short, that integrates the quantum computing paradigm with the paradigm of quantum photonics. We analyzed and benchmarked this model against a number of alternative models using the standard MNIST dataset.

Extending on the prior work on quantum machine learning frameworks (\hyperref[c10]{\cite{c10}}\hyperref[c11]{\cite{c11}}\hyperref[c17]{\cite{c17}}\hyperref[c18]{\cite{c18}}\hyperref[c19]{\cite{c19}}\hyperref[c25]{\cite{c25}}), our QOCNN architecture combines a quantum convolution layer to extract salient information from the input image data with an efficient and surprisingly robust quantum optical neural network, which is a quantum-optical replacement for the multilayer perceptron.

We have shown that the proposed QOCNN model achieves accuracies that are similar to the benchmark models, outperforms them in robustness, while also having significant theoretical potential for outperforming its peers in computational efficiencies when executed on the future quantum computing hardware.

All in all, the overall effectiveness of the quantum optical convolutional neural network shows the overwhelming potential that quantum computing contains for the future of artificial intelligence and machine learning.

With the rapid pace of progress in the research and advancement in the fields of quantum computers and quantum photonics with fault-tolerant NISQ qubits, and the increasingly more effective quantum computers being developed and reported by many organizations, this novel framework could enable substantially higher computing capabilities than the current classical machine learning frameworks, and help unleash new frontiers of AI applications.

\addtolength  {\textheight}{-18.4cm}   



\section*{\textbf{APPENDIX}}

The aforementioned GitHub by Mike Fang that this paper used as a basis is \href{https://github.com/mike-fang/imprecise_optical_neural_network}{\underline{\color{blue}{here}}} \hyperref[c18]{\cite{c18}}.

To see the above code and our additions, modifications, and data, go \href{https://github.com/rishab-partha/Quantum-Optical-ConvNet}{\underline{\color{blue}{here}}}.

\section*{\textbf{ACKNOWLEDGMENT}}

We would like to thank the advisors of the Research Program at The Harker School, Ms. A. Chetty and Dr. C. Spenner, for affording us the opportunity to conduct this research. Next, we would like to thank our faculty advisor at Harker, Dr. E. Nelson, who helped us form and shape our research ideas. We would also like to thank our parents for their unwavering support.


\end{document}